# Detecting Phone-Induced Pedestrian Distraction via a Multimodal Fusion Transformer

Yuanzhe Li
*Chair of Automotive Engineering*
*Technische Universität Berlin*
Berlin, Germany

Hounian Liu
*Chair of Automotive Engineering*
*Technische Universität Berlin*
Berlin, Germany

Xiaotong Chang
*Chair of Automotive Engineering*
*Technische Universität Berlin*
Berlin, Germany

Yidi Huang
*Chair of Automotive Engineering*
*Technische Universität Berlin*
Berlin, Germany

***Abstract*—The increasing reliance on mobile phones has made phone-induced pedestrian distraction increasingly prevalent. Activities such as texting, watching videos, and making phone calls have become significant contributors to traffic accidents. Reliable detection of pedestrian distraction is essential for autonomous vehicles, as it improves situational awareness and enables timely risk assessment, thereby supporting safe motion planning and vehicle control. We propose a multimodal fusion Transformer (MFT) for detecting phone-induced pedestrian distraction. MFT jointly extracts skeletal dynamics from body pose keypoints and visual appearance features from pedestrian images, effectively leveraging the complementary information provided by the two modalities. A cross-modal attention module is proposed to capture inter-modal dependencies through multi-head cross-attention, facilitating effective fusion of complementary information across the two modalities. Then, a temporal attention fusion module, implemented with a Transformer encoder, is employed to capture temporal dependencies. MFT is trained and evaluated on a manually annotated dataset comprising 287 pedestrian instances with 20,741 images. Extensive experiments demonstrate that MFT attains an overall accuracy of 95%, exceeding the performance of six baseline approaches by 6%.**

## I. Introduction

Pedestrian-vehicle crashes have increased significantly in recent years. In the United States, traffic accidents in 2023 resulted in 7,314 pedestrian fatalities and over 68,000 injuries [22]. Pedestrian distraction is one of the major factors contributing to pedestrian-vehicle crashes. With the rapid proliferation of mobile phones and their increasing importance in daily life, the primary source of pedestrian distraction arises from mobile phone use, including texting, video viewing, navigation, gaming, and phone conversations. As reported in [2], pedestrian injuries associated with mobile phone use have increased more than twofold since 2005.

The rapid development of autonomous vehicles (AVs) is expected to improve traffic efficiency and enhance road safety[1, 3]. Accurate detection of phone-induced pedestrian distraction is essential for AVs, as it provides critical pedestrian-related information that serves as an important cue for risk assessment and further facilitates safe decision-making and vehicle control. However, phone-induced pedestrian distraction is difficult to detect because mobile phones are small and are often occluded by the pedestrian's body or nearby objects, reducing their visibility in video data and complicating the accurate recognition of distraction behaviors.

Phone-induced pedestrian distraction detection has emerged as an important research topic. Early approaches mainly relied on static single-frame inputs. For instance, convolutional neural networks (CNNs) have been used to capture visual features from pedestrian images [6] or motion cues from pose keypoints [7]. Rangesh et al. [4] proposed a single-frame vision-based framework that combines articulated human pose estimation and local hand-region features, with a belief network fusing pose clusters and mobile-phone presence cues to estimate pedestrian distraction probability. In [5], the authors introduced a multicue framework that leveraged a single pedestrian image and integrated articulated pose estimation, hand-object interaction analysis, and gaze direction modeling to achieve accurate and computationally efficient detection of phone-induced pedestrian activities.

In subsequent studies, synchronized image pairs captured by a dual-camera system, as well as multi-frame sequences, have been employed as inputs to exploit information across different viewpoints and temporal dimensions. For instance, in [8], a two-branch CNN is proposed, which extracts features from pedestrian images captured by the left and right camera views separately, and detects pedestrian distraction behaviors through feature concatenation and fully connected layers. The authors further showed that extending from single-frame to image-pair and short video inputs, with confidence accumulation and voting, improves detection stability. In [9], PPDBNet is proposed, which employs separate long short-term memory (LSTM) networks to model the temporal dynamics of fused image and pose features derived from dual-view inputs, based on sequences of 10 consecutive frames.

Building upon these studies, we focus on detecting pedestrian distraction from single-view image pairs, aiming to effectively capture temporal cues while classifying each pedestrian as eye engagement, call engagement, or no engagement. A multimodal fusion Transformer is designed to extract skeletal dynamics from body pose keypoints and visual appearance features from pedestrian RGB images. Motivated by the remarkable success of Transformers in both computer vision [15] and sequential modeling tasks [11, 21], we adopt Transformer and its variants for extracting spatial representations and modeling temporal information. The main contributions are as follows: (1) We propose a multimodal fusion Transformer (MFT) for detecting phone-induced pedestrian distraction, which mines complementary cues from pedestrian images and body pose keypoints. (2) A hierarchical fusion strategy is employed, in which a cross-modal attention module captures inter-modal interactions to facilitate complementary feature fusion, followed by a temporal attention fusion module that models temporal dependencies. (3) Comprehensive experimental evaluations and ablation analyses validate the performance of MFT and support the design choices underlying its architecture.

## II. Methodology

### A. Problem Formulation

Phone-induced pedestrian distraction detection is addressed as a three-class classification task. Given a fixed-length observation sequence, the goal is to classify the target pedestrian into one of three distraction categories: (1) Eye engagement, which includes phone-related activities including video viewing, gaming, text messaging, and map navigation. (2) Call engagement, corresponding to making or receiving phone calls. (3) No engagement, indicating that the pedestrian is not involved in any phone-related activity. The observation window is defined as $N$=5 consecutive frames.

MFT employs two dedicated encoders to separately extract visual appearance features from pedestrian images and skeletal dynamics from body pose keypoints, which are subsequently integrated through cross-modal attention and temporal attention fusion. The overall architecture of MFT is shown in Fig. 1, and its core modules are introduced in the following sections.

### B. Visual Appearance Encoder

The visual appearance encoder extracts visual appearance features from pedestrian images that contain rich visual cues, such as appearance texture, body posture, limb movements, hand gestures, and mobile phone appearance. These cues reflect the pedestrian's behavioral state and attention distribution, which are of great importance for detecting pedestrian distraction. Considering the superior performance of Transformers over traditional convolutional neural networks in vision tasks, Swin Transformer V2 [19] is adopted as the backbone to extract visual features. Swin Transformer V2 benefits from the hierarchical design with shifted windows, enabling efficient computation and strong capability in modeling both local and global visual contexts. To balance predictive performance and computational cost, we employ a Swin V2-Base backbone pretrained on ImageNet [16]. The pedestrian crops are resized to 256×256 before being fed into the backbone. The spatial feature maps from the final layer are extracted, globally average-pooled, and then projected by a multilayer perceptron (MLP) from 1024 to 128 to reduce computation, producing the visual feature sequence $F_V \in \mathbb{R}^{N\times 128}$.

### C. Skeletal Dynamics Encoder

The skeletal dynamics encoder extracts motion-related skeletal cues from pedestrian body pose keypoints. The body pose keypoints contain rich kinematic and structural information that reflects the body's motion patterns and posture dynamics, providing cues for behaviors such as looking at a phone, and making a phone call. The pretrained HRNet model [12] is used to extract the body pose keypoints $P = \{ p^{t-N+1}, p^{t-N+2}, ..., p^{t} \} \in \mathbb{R}^{N\times 36}$ from the pedestrian images, where $p^t$ denotes the pose keypoints of the $t$-th frame, which include the ($x$, $y$) coordinates of body joints, such as the head, wrists, etc. Note that HRNet does not provide the neck keypoint, we synthesize the neck as the midpoint of the left and right shoulders. This yields an 18-joint representation for each frame:

$$p^t = \{ x_1^t, y_1^t, x_2^t, y_2^t, ..., x_{18}^t, y_{18}^t \} \quad (1)$$

The pose keypoints $P$ are embedded into a high-dimensional representation, with positional encodings added to retain temporal ordering. To model the temporal and structural dependencies among pose keypoints, a Transformer encoder [17] is employed to extract high-level feature representations. The Transformer encoder comprises multiple layers, each composed of multi-head self-attention and feed-forward networks, which capture structural representations and temporal dependencies among pose keypoints, and produce the final pose feature sequence $F_P \in \mathbb{R}^{N\times 128}$.

### D. Cross-Modal Attention

To effectively fuse the visual and pose features, a cross-modal attention module is introduced to enable frame-wise interactions between the two modalities. Specifically, for each modality $i$, where $i \in \mathcal{M} = \{P, V\}$, a cross-attention mechanism with $n$ parallel attention heads is applied, where query is from modality $i$, while the keys and values are obtained from both modalities. In this way, each modality interacts with both itself and the other modality to refine its feature representation. Let $f_i^\tau$ denote the features of modality $i$ at time step $\tau$. For the $k$-th attention head, the attention weight from modality $i$ to modality $j$, where $i, j \in \mathcal{M}$, is calculated as follows:

$$\alpha_{i,j}^k = \frac{\exp\left(\frac{Q_i^k \cdot K_j^k}{\sqrt{d}}\right)}{\sum_{s=1}^{2} \exp\left(\frac{Q_i^k \cdot K_s^k}{\sqrt{d}}\right)} \quad (2)$$

where $Q_i^k = f_i^\tau W_k^Q$ and $K_j^k = f_j^\tau W_k^K$ are query and key of modality $i$ and modality $j$, respectively.

The attention weight is employed to guide information exchange between the two modalities:

$$CA_k = \sum_{j\in\mathcal{M}} \alpha_{i,j}^k V_j^k \quad (3)$$

where $V_j^k = f_j^\tau W_k^V$ is the value of modality $j$.

The refined features of modality $i$ at time step $\tau$ is refined:

$$\hat{f}_i^\tau = [CA_1, CA_2, ..., CA_n] \cdot W_i^O \quad (4)$$

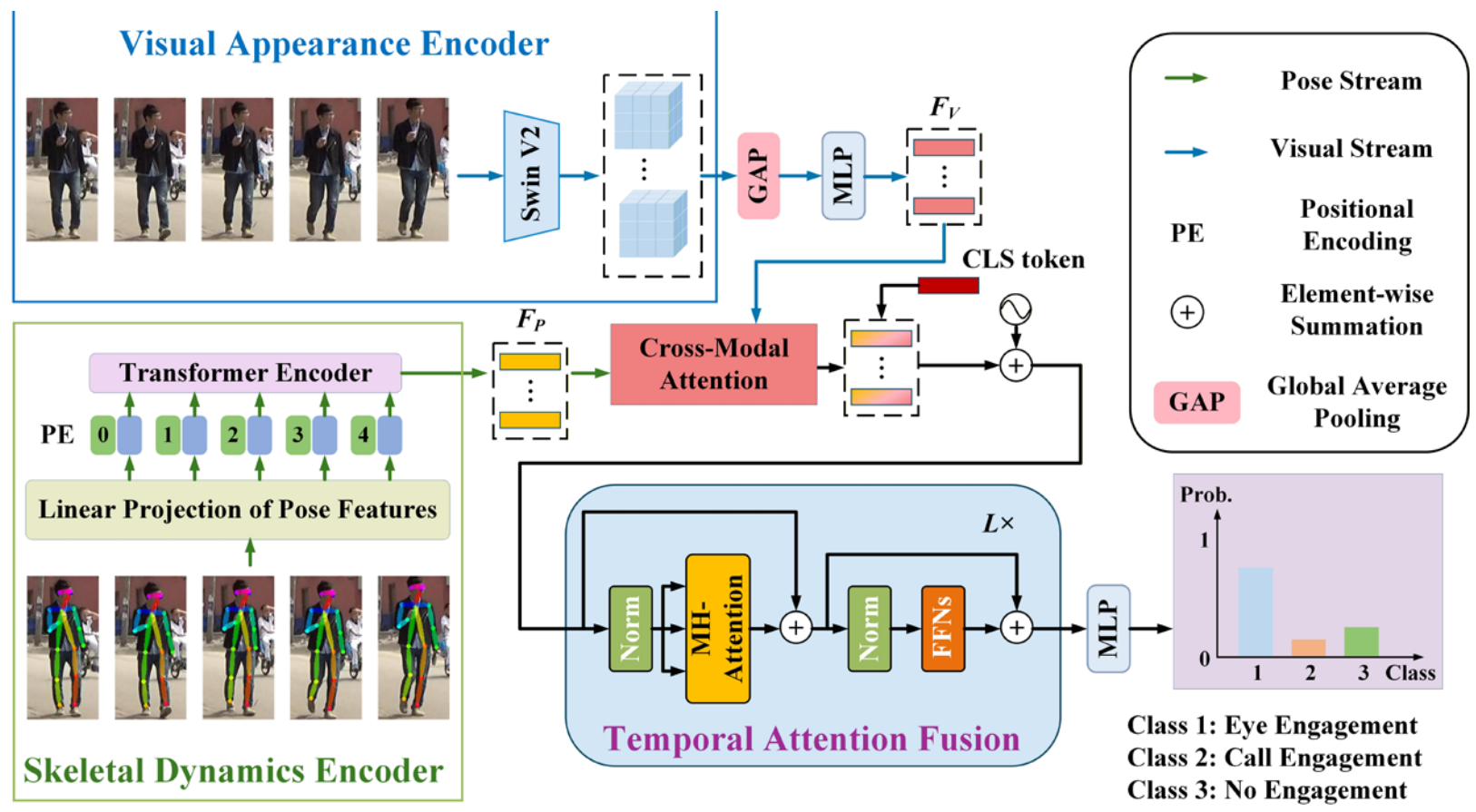

Fig.1. The overall architecture of the proposed multimodal fusion Transformer.

Through the above process, the refined feature sequences $\hat{F}_P \in \mathbb{R}^{N\times128}$ and $\hat{F}_V \in \mathbb{R}^{N\times128}$ are obtained. The fused feature $F_f \in \mathbb{R}^{N\times128}$ is obtained by frame-wise averaging the corresponding features in $\hat{F}_P$ and $\hat{F}_V$.

### E. Temporal Attention Fusion

The temporal attention fusion module further integrates the fused features $F_f$ along the temporal dimension, where a Transformer encoder [17] is applied to model the temporal dependencies. Specifically, a CLS token is prepended to the fused features $F_f$ to form a global representation of the entire sequence. To compensate for the lack of temporal positional awareness, sinusoidal positional encoding is added to the sequence. The resulting feature sequence $F_f' \in \mathbb{R}^{(N+1)\times128}$ is then fed into a Transformer encoder consisting of multiple layers, each containing a multi-head self-attention (MHSA) and a feed-forward network (FFN) sublayer. In each sublayer, the input is first normalized, and a residual connection is employed to integrate the sublayer output with the original input. The output of the $l$-th Transformer layer can be expressed as follows:

$$T_f^l = \text{FFN}\left(\text{Norm}\left(\hat{T}_f^l\right)\right) + \hat{T}_f^l \tag{5}$$

$$\hat{T}_f^l = \text{MHSA}\left(\text{Norm}\left(T_f^{l-1}\right)\right) + T_f^{l-1} \tag{6}$$

where $T_f^l$ and $\hat{T}_f^l$ is the output of FFN and MHSA sub layer, respectively. Norm(·) denotes the layer normalization.

The MHSA sublayer contains $m$ attention heads operating in parallel. For the $h$-th attention head, the query $Q_h$, key $K_h$, and value $V_h$ are obtained through learnable linear transformation $W_h^Q$, $W_h^K$, $W_h^V$ to the input feature sequence. The temporal attention matrix is computed as:

$$\alpha_h = \text{Softmax}\left(\frac{Q_h \cdot K_h^T}{\sqrt{d}}\right) \tag{7}$$

where $\alpha_h \in \mathbb{R}^{(N+1)\times(N+1)}$ is the temporal attention matrix that reflects how each frame attends to the others.

By multiplying $V_h$ with $\alpha_h$, the CLS token integrates information from the entire sequence, while the feature representation of each frame is refined by incorporating contextual cues from other frames across the sequence.

$$SA_h(Q_h, K_h, V_h) = \alpha_h \cdot V_h \tag{8}$$

The outputs from all self-attention heads are concatenated and linearly transformed using a learnable projection matrix:

$$MHSA(Q,K,V) = \left[SA_1, SA_2, \ldots, SA_m\right] \cdot W^O \tag{9}$$

The resulting features are further refined by the FFN sublayer. The final-layer CLS token, which summarizes the temporal sequence, is then passed to an MLP for distraction classification, with softmax producing the probabilities for the three classes.

## III. Experiments

### A. Dataset and Experimental Settings

We conducted the experiment using images from the Labeled Pedestrians in the Wild dataset [18]. We manually annotated a total of 287 pedestrians, including 106 eye engagement, 59 call engagement, and 122 no engagement instances, resulting in 20,741 annotated images in total. The details of the annotated dataset are summarized in Table I. Illustrative examples of the annotated images and their corresponding pose keypoints are presented in Fig. 2.

The dataset is split into 201, 43, and 43 pedestrians for training, testing, and validation, respectively. The observation length is set to 5 frames, and the sliding window moves with a stride of 2. This setting generates 6,961, 1,425, and 1,182 sequences for training, testing, and validation.

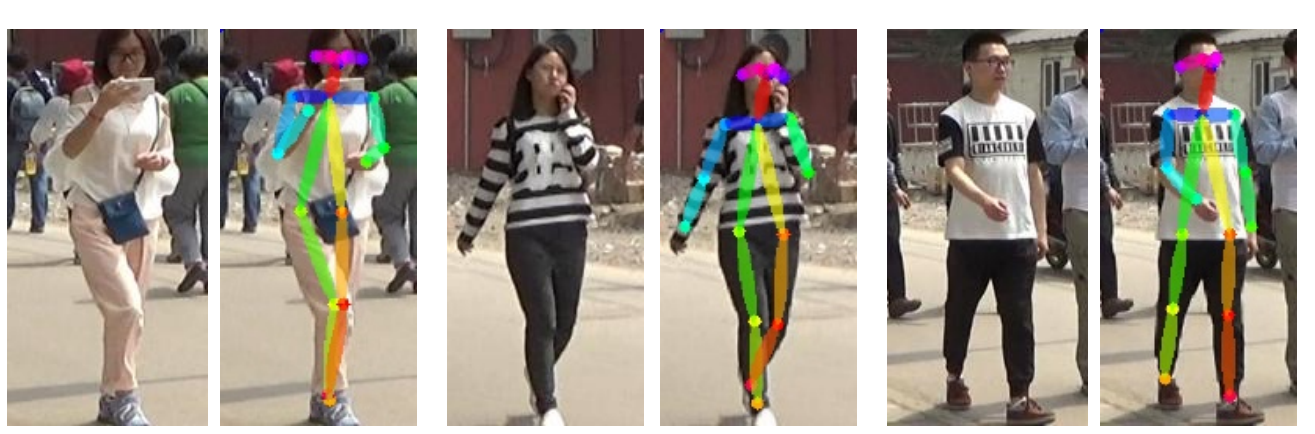

(a) Eye engagement (b) Call engagement (c) No engagement

Fig. 2. Examples of the annotated data.

TABLE I. SUMMARY OF THE ANNOTATED DATASET

| Class | No. of Pedestrians | No. of Images |
|---|---|---|
| Eye engagement | 106 | 8,116 |
| Call engagement | 59 | 4,658 |
| No engagement | 122 | 7,967 |
| Total | 287 | 20,741 |

The Transformer encoder is configured with a hidden dimension of 128, 4 attention heads, and 2 layers for skeletal dynamics encoder and the temporal attention fusion module. The classification head comprises two fully connected layers with 128 hidden units and three output units, respectively. To mitigate overfitting, a dropout with a rate of 0.3 is applied after the first hidden layer. We employ the Focal Loss as loss function, where the class-balancing factor $\alpha$ is set according to the class distribution weights, and the focusing parameter $\gamma$ is fixed to 2. The model is optimized using Adam optimizer with a learning rate of $1\times10^{-5}$, and trained for 200 epochs with a batch size of 8.

### *B. Evaluating Metrics*

MFT is evaluated using precision (P), recall, F1-score, and overall accuracy (Acc). Precision denotes the fraction of correctly predicted samples among all samples assigned to a specific class, reflecting the reliability of the model's class predictions. Recall describes the fraction of ground-truth instances that are successfully recognized, indicating the model's ability to capture positive samples. The F1-score combines precision and recall through their harmonic mean, offering a balanced assessment of classification performance. Overall accuracy quantifies the percentage of samples correctly predicted across the three distraction classes, reflecting the model's general classification capability.

### *C. Comparison Methods*

To validate the effectiveness of the proposed MFT, we compare it with several baseline methods. The first group of baselines is based on static single-frame input, with each method operating on individual frame independently, without incorporating temporal information. These include: (1) Deep ConvNet [6]: A deep convolutional neural network architecture based on pedestrian image for pedestrian distraction detection. (2) MobileNet [14]: A MobileNet [13] pretrained on ImageNet dataset [16] is used to extract visual features of pedestrian. (3) PoseCNN [7]: The body pose keypoints are used as input and processed by a CNN for classification.

The second group of baselines involves temporal modeling, including (1) VGG16-S: We reproduce a variant of Two-Branch CNN [8], where a single-view image sequence is used. VGG16 [10] is adopted as the visual backbone to extract features from consecutive frames, and the voting mechanism is applied in the same manner as Two-Branch CNN [8]. (2) AlexNet-S: The same architecture as VGG16-S, except that VGG16 is replaced with AlexNet [20] as the backbone. (3) PPDBNet-S: We reproduce a single-view variant of PPDBNet [9]. A 5-frame sequence and the corresponding poses are encoded by VGG16, followed by an LSTM and temporal pooling for classification.

### *D. Comparative Experimental Results*

Table II compares MFT with the baseline methods, with the best, second-best, and third-best results highlighted in green, blue, and yellow, respectively. The results demonstrate that the proposed MFT achieves the highest overall accuracy of 95%, surpassing the second-best methods, PoseCNN [7] and VGG16-S [8], which both reach 89%. Compared with the static single-frame based methods such as Deep ConvNet [6] and MobileNet [14], MFT exhibits substantial performance gains. For instance, Deep ConvNet achieves only 70% overall accuracy, while MobileNet improves to 84%, yet both remain notably below MFT. Among the first group of methods, PoseCNN [7] achieves the best performance with an overall accuracy of 89%, benefit-ing from the use of body pose information that provides richer cues than purely visual features. However, it still lags behind MFT. This demonstrates the advantages of MFT in both multimodal fusion, which effectively integrates complementary information across modalities rather than relying on a single modality alone, and temporal modeling, which captures dynamic behavioral cues over time. Among the second group of methods, MFT also consistently outperforms all baselines. Specifically, VGG16-S [8] achieves an overall accuracy of 89%, while AlexNet-S [8] and PPDBNet-S [9] reach 81% and 88%, respectively. MFT surpasses these three baselines by a significant margin, demonstrating its superior capability to effectively integrate visual and pose information through cross-modal attention and temporal attention fusion mechanisms.

TABLE II. COMPARATIVE RESULTS WITH BASELINE METHODS

| Model | Class 1 | | | Class 2 | | | Class 3 | | | Overall |
|---|---|---|---|---|---|---|---|---|---|---|
| | P | Recall | F1 | P | Recall | F1 | P | Recall | F1 | Acc |
| Deep ConvNet [6] | 0.75 | 0.77 | 0.76 | 0.37 | 0.46 | 0.41 | 0.77 | 0.70 | 0.73 | 0.70 |
| MobileNet [14] | 0.91 | 0.79 | 0.85 | 0.61 | 0.75 | 0.67 | 0.86 | 0.91 | 0.88 | 0.84 |
| PoseCNN [7] | **0.94** | 0.82 | **0.88** | 0.66 | **0.98** | **0.79** | **0.94** | **0.92** | **0.93** | **0.89** |
| VGG16-S [8] | 0.88 | **0.91** | **0.89** | **0.86** | **0.88** | **0.87** | 0.90 | 0.87 | 0.89 | **0.89** |
| AlexNet-S [8] | 0.82 | 0.84 | 0.83 | **0.69** | 0.45 | 0.55 | 0.83 | 0.89 | 0.86 | 0.81 |
| PPDBNet-S [9] | **0.92** | **0.86** | **0.89** | 0.65 | 0.84 | 0.73 | **0.94** | **0.92** | **0.93** | 0.88 |
| **MFT** | **0.95** | **0.97** | **0.96** | **0.84** | **0.97** | **0.90** | **0.98** | **0.93** | **0.95** | **0.95** |

TABLE III. ABLATION STUDY RESULTS

| Model | Class 1 | | | Class 2 | | | Class 3 | | | Overall |
|---|---|---|---|---|---|---|---|---|---|---|
| | P | Recall | F1 | P | Recall | F1 | P | Recall | F1 | Acc |
| MFT-v1 | 0.89 | 0.90 | 0.90 | 0.75 | **0.95** | 0.84 | 0.93 | 0.86 | 0.89 | 0.89 |
| MFT-v2 | 0.80 | 0.89 | 0.84 | 0.76 | 0.58 | 0.66 | 0.92 | 0.88 | 0.90 | 0.85 |
| MFT-v3 | 0.92 | **0.95** | **0.94** | 0.71 | 0.85 | 0.77 | **0.97** | 0.88 | 0.92 | 0.91 |
| MFT-v4 | 0.92 | 0.95 | 0.93 | 0.76 | 0.86 | 0.81 | **0.97** | 0.91 | 0.94 | **0.92** |
| MFT-v5 | 0.88 | 0.91 | 0.90 | 0.74 | 0.71 | 0.72 | 0.96 | **0.94** | **0.95** | 0.90 |
| MFT-v6 | **0.95** | 0.87 | 0.91 | **0.80** | 0.94 | **0.87** | 0.92 | **0.95** | 0.94 | 0.91 |
| **MFT** | **0.95** | **0.97** | **0.96** | **0.84** | **0.97** | **0.90** | **0.98** | 0.93 | **0.95** | **0.95** |

In terms of class-level performance, MFT achieves competitive results. For class 1 (eye engagement), MFT attains the highest precision, recall, and F1-score, reaching 95%, 97%, and 96%, respectively. These results surpass the second-best method PoseCNN [7] by 1 percentage points (pp) in precision, and exceed VGG16-S [8] by 6 pp in recall and 7 pp in F1-score. For class 2 (call engagement), MFT ranks second in both precision and recall, achieving 84% and 97%, respectively. It is slightly lower than VGG16-S [8] by 2 pp in precision and PoseCNN [7] by 1 pp in recall. Nevertheless, MFT attains the highest F1-score of 90%, demonstrating a well-balanced performance between precision and recall. For class 3 (no engagement), MFT achieves the highest precision, recall, and F1-score of 98%, 93%, and 95%, respectively. These results surpass the second-best methods PPDBNet-S [9] and PoseCNN [7] by 4, 1, and 2 pp, respectively.

Experimental results demonstrate that MFT can accurately detect phone-induced pedestrian distractions. Its superior performance stems from the multimodal fusion architecture, which effectively extracts complementary information from both pedestrian image and body pose keypoints. In addition, the dense cross-modal interactions enable more comprehensive feature fusion, while the efficient Transformer-based temporal modeling further enhances the network's ability to capture temporal dependencies.

### *E. Qualitative Results*

We provide qualitative results to illustrate the performance of MFT in typical scenarios, as shown in Fig. 3. The left side of each case shows a sequence of five pedestrian image frames along with their corresponding poses, while the right side presents the estimated probability for each class. In case (a), a man is holding a phone in one hand while swinging the other arm as he walks, with his gaze directed forward. MFT correctly identifies a 94.63% probability for the call engagement. In case (b), a man is holding a phone in one hand while walking, with the other hand holding a book against his chest instead of swinging naturally. MFT identifies a 99.83% probability for call engagement, accurately identifying the behavior rather than misclassifying it as "looking at the phone." In case (c), a man is holding a phone in one hand and focusing his gaze on the screen. MFT classifies this behavior as eye engagement with a 99.84% probability. The qualitative results further validate that MFT maintains stable performance and effectively integrates complementary cues from visual and pose features, leading to accurate pedestrian distraction detection.

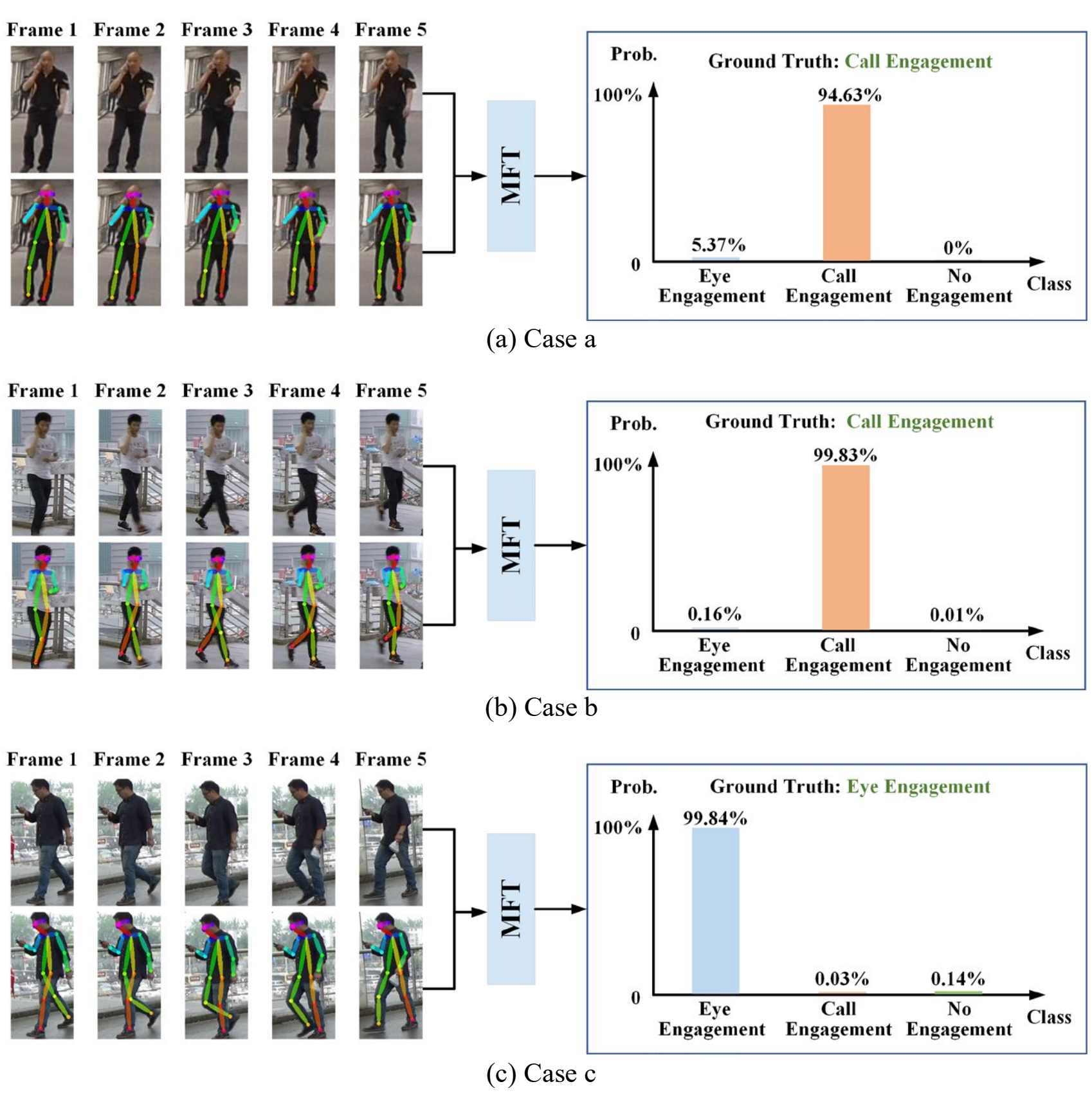

Fig. 3. Qualitative results

### F. Ablation Study

To assess the contribution of each modality and architectural component, we conduct an ablation study with several variants: (1) MFT-v1 removes the visual appearance encoder. (2) MFT-v2 removes the skeletal dynamics encoder. (3) MFT-v3 replaces Swin Transformer v2 with VGG16. (4) MFT-v4 replaces the Transformer-based temporal fusion with LSTM. (5) MFT-v5 uses GRU for temporal fusion. (6) MFT-v6 replaces cross-modal attention with element-wise addition. Table III reports the results, with the best and second-best values highlighted in green and blue, respectively.

Compared with MFT, both MFT-v1 and MFT-v2 show performance degradation, with overall accuracy drops of 6 and 10 pp, respectively. In particular, the detection of call engagement declines noticeably. When the visual encoder is removed, precision for call engagement drops from 84% to 75%, indicating more false positives. When the pose encoder is removed, recall drops sharply from 97% to 58%, showing that pose cues play a crucial role in accurately detecting the call engagement. The results indicate that the proposed multimodal fusion framework effectively exploits complementary information from visual and pose cues, yielding superior overall performance over single-modality variants.

The performance of MFT-v3 decreases, with the overall accuracy dropping by 4 pp, indicating that Swin Transformer v2 serves as a more effective visual backbone than traditional CNNs by capturing richer and more discriminative visual features. Compared with MFT, MFT-v4 and MFT-v5 also show performance drops, with overall accuracy decreases of 3 and 5 pp, respectively. This demonstrates that Transformer encoder is more effective than LSTM and GRU in capturing temporal dependencies. The performance degradation of MFT-v6 is attributed to the fact that all modalities are treated equally, without capturing their relative contributions. In contrast, the cross-modal attention module in MFT enables frame-wise interaction between the two modalities, allowing the network to adaptively model their complementary relationships and assign attention weights more effectively.

## IV. Conclusion

This study presents a multimodal fusion Transformer for detecting phone-induced pedestrian distraction. The proposed network employs two dedicated encoders to extract skeletal dynamics from body pose keypoints and visual appearance features from pedestrian images, thereby exploiting the complementary strengths of the two modalities. A cross-modal attention module is introduced to facilitate the integration of visual and pose features by capturing inter-modal interactions, while a Transformer-based temporal attention fusion module is introduced to capture dependencies across the sequence. Comparative evaluations confirm the effectiveness of the proposed network, which attains an overall accuracy of 95% and consistently surpasses the baseline approaches. Future research will investigate more efficient model designs and explore practical deployment in real-world applications.